\documentclass{clv3}

\usepackage{multicol}
\usepackage{tipa}
\usepackage{longtable}
\usepackage{booktabs}
\usepackage{hyperref}
\usepackage{xcolor}
\usepackage{layouts}
\definecolor{darkblue}{rgb}{0, 0, 0.5}
\hypersetup{colorlinks=true,citecolor=darkblue, linkcolor=darkblue, urlcolor=darkblue}

\begin{document}


\runningtitle{Perception of Phonological Assimilation}

\runningauthor{Pouw et al.}


\title{Perception of Phonological Assimilation by Neural Speech Recognition Models}


\author{Charlotte Pouw}
\affil{ILLC, University of Amsterdam}

\author{Marianne de Heer Kloots}
\affil{ILLC, University of Amsterdam}

\author{Afra Alishahi}
\affil{CSAI, Tilburg University}

\author{Willem Zuidema}
\affil{ILLC, University of Amsterdam}

\maketitle

\begin{abstract}
Human listeners effortlessly compensate for phonological changes during speech perception, often unconsciously inferring the intended sounds. For example, listeners infer the underlying /n/ when hearing an utterance such as "clea[m] pan", where [m] arises from place assimilation to the following labial [p]. This article explores how the neural speech recognition model Wav2Vec2 perceives assimilated sounds, and identifies the linguistic knowledge that is implemented by the model to compensate for assimilation during Automatic Speech Recognition (ASR). Using psycholinguistic stimuli, we systematically analyze how various linguistic context cues influence compensation patterns in the model's output. Complementing these behavioral experiments, our probing experiments indicate that the model shifts its interpretation of assimilated sounds from their acoustic form to their underlying form in its final layers. Finally, our causal intervention experiments suggest that the model relies on minimal phonological context cues to accomplish this shift. These findings represent a step towards better understanding the similarities and differences in phonological processing between neural ASR models and humans.


\end{abstract}


\section{Introduction}

When listening to spoken language, the human perceptual system is faced with a challenge: speech segments can surface in many different ways,  depending on factors such as their phonological context and the unique characteristics of the speaker producing them. Any speech recognition system must learn to recognize the intended words regardless of the various ways in which those words may be pronounced.

A substantial amount of the variability in speech is systematic, arising from phonological processes occurring in predictable environments. One such process is \textbf{place assimilation}, where phonemes adopt the articulation place of adjacent phonemes. For instance, the word pair \textit{clean pan} is frequently pronounced as \textit{clea[m] pan}, with the word-final coronal /n/ in \textit{clean} assimilating to the subsequent labial [p] in \textit{pan}. This is a simple yet common phonological process across the world's languages \cite{hura1992role}. In English, it occurs for coronal
segments (e.g., /t/, /d/, /n/) that are followed by noncoronals,
such as labials (e.g., [p], [b], [m]) or velars (e.g., [k], [g], [\textipa{N}]).

Human listeners are able to infer the underlying /n/ when exposed to assimilated inputs like \textit{clea[m] pan}, allowing them to perceive the intended word \textit{clean}. This phenomenon is referred to as \textbf{compensation} for assimilation and happens automatically---that is, humans compensate without conscious awareness of the assimilation itself. 
Psycholinguistic research has used controlled stimuli to investigate the mechanism behind this process. Several experimental paradigms have been used, including cross-modal priming \cite{gaskell1996phonological}, phoneme monitoring \cite{weber2001help}, and word detection \cite{darcy2009phonological}. Results from these studies have lead to clear insights about the linguistic cues that facilitate human assimilation processing.



It is currently an open question how computational models for speech recognition process assimilated sounds. Traditionally, Automatic Speech Recognition (ASR) systems accounted for phonological changes using a \textit{pronunciation dictionary}, which explicitly stored multiple pronunciations of each lexical item. However, in recent years, the state-of-the-art has become defined by end-to-end systems based on neural architectures that are trained in a self-supervised fashion. In learning to map directly from raw acoustic signals to text transcriptions, such systems must find implicit ways to account for phonological changes. 
Given their learning setup, these models are "black boxes"---hence it remains unclear how exactly they address challenges such as place assimilation.

There is some evidence that self-supervised speech models implicitly encode sophisticated linguistic knowledge, including phonological, lexical and even syntactic information \cite{pasad2021layer,pasad2023self,shen2023wave}, but these findings are predominantly correlation-based and lack a direct connection to downstream behavior in tasks such as ASR. In contrast, psycholinguistic stimuli are designed to require a dependence on precise linguistic knowledge to achieve specific behavioral responses. Insights and materials from psycholinguistic research thus offer an invaluable resource for systematically analyzing the linguistic knowledge encoded by self-supervised speech models. 

In this study, we draw inspiration from psycholinguistic research to investigate the extent to which neural ASR models compensate for place assimilation, and which linguistic cues allow them to do so. We start by analyzing the \textbf{behavior} of such models when exposed to controlled psycholinguistic stimuli. We then perform a series of \textbf{interpretability} experiments to better understand the mechanism behind the observed behavior. Our objective is twofold. Firstly, we aim to assess the extent to which the output of neural ASR models follows human-like patterns in compensating for place assimilation. Secondly, we aim to shed light on the specific linguistic knowledge encoded by neural ASR models and to determine the causal role of that knowledge in accounting for place assimilation. 



\section{Related Work}\label{sec:theories}

Our study of compensation for assimilation in neural ASR models is grounded in a large body of work studying the phenomenon in humans through psycholinguistic experiments and computational modelling. Here, we review theories and experimental findings regarding the \emph{levels of linguistic knowledge} involved in compensating for place assimilation in humans and traditional ASR systems.

We do not review advances in neural speech recognition, nor advances in the analysis and interpretation of deep learning models. We assume both topics are well-known in the computational linguistics community; for more background, we refer to \citet{li2022recent} for a survey of recent automatic speech recognition models, to \citet{lyu2024towards} for a survey of interpretability techniques in NLP, and to \citet{alishahi2017encoding} and \citet{giulianelli2018under} for early examples of interpretability combined with speech models or causal interventions, respectively.

\subsection{Compensation in Humans}\label{sec:background}

Human speech perception is known to be informed by learned linguistic knowledge on the level of phonemes and on the level of words. Both levels may influence the perception of assimilated speech sounds, resulting in the two complementary processes of \emph{phonological} and \emph{lexical compensation}.

\textbf{Lexical compensation} proposes that listeners use a stored list of lexical items to match the incoming signal against. This mechanism essentially treats place assimilation as random noise and also compensates for irregular changes (i.e., those not attributable to phonological processes) such as mispronunciations. \citet{lahiri1991mental} argue that mental representations of lexical items are underspecified; that is, not all phonetic features of the lexical items are explicitly stored. This underspecification allows for a single mental representation to accommodate a range of phonological variations that speakers may produce in different contexts. An important prediction of this line of theories is that compensation is \textbf{independent} of phonological context; that is, listeners should be able to derive the intended word regardless of the phonological environment in which the change occurs. Higher-order semantic cues may be used to infer which word is likely in the context.

\textbf{Phonological compensation} (or phonological \textit{inference}, \citealp{GASKELL1995407}) proposes that listeners use knowledge of phonological rules to infer the underlying form of an altered segment. In this line of theories, the same mechanism that handles regular phonological processes such as elision or insertion is used to compensate for place assimilation. This mechanism is \textbf{sensitive} to phonological context: it checks whether the assimilation occurs in a phonologically viable environment, and uses this information to infer the underlying form. Unlike lexical compensation, it operates without integrating semantic cues.

Psycholinguistic research supports the idea that human listeners are sensitive to phonological context when compensating for place assimilation. \citet{gaskell1996phonological} find that humans tend to perceive assimilated forms such as \textit{wicke[b]} as the underlying form \textit{wicked} if the assimilation is viable within the phonological context. That is, humans responded faster in a lexical decision task when they heard sequences like \textit{wicke[b] prank} as compared to \textit{wicke[b] game}, because the labial change /d/ $\rightarrow$ [b] is only triggered by [p] in \textit{prank}, not by [g] in \textit{game}. This finding is robust across experimental paradigms \cite{gaskell1998mechanisms} and across languages \citep{coenen2001variation,mitterer2003coping,mitterer2006role,darcy2009phonological}, providing evidence that compensation for assimilation in humans cannot be explained by a purely lexical mechanism.

Nevertheless, there is some evidence that humans use semantic context to process specific cases of assimilation. In English, place assimilation can be complete, leaving no detectable acoustic traces of the underlying phoneme \citep{nolan1992descriptive,ellis2002categorical,dilley2007study}. When assimilation is incomplete, listeners can leverage traces of the original phoneme to infer the intended word \cite{gow2002does}. However, complete place assimilation can lead to lexical ambiguity, exemplified in \textit{I think a quick ru[m] picks you up}, where \textit{ru[m]} could be either the standard pronunciation of \textit{rum} or the assimilated form of \textit{run}. This illustrates the neutralization of a phonemic contrast, where there is no difference in the surface realization of the phonemes /n/ and /m/. \citet{gaskell2001lexical} find that individuals tend to perceive the surface form \textit{rum} in neutral sentential contexts, indicating that acoustic evidence has priority. However, introducing a biasing sentential context (e.g., \textit{"It’s best to start the day with a burst of activity"}) activates both lexical candidates (\textit{run} and \textit{rum}), suggesting that human listeners are helped by semantic cues in resolving the lexical ambiguity that results from complete place assimilation.

\subsection{Compensation in ASR models}

The minimal context cues leading to human-like compensation for assimilation have also been explored using computational systems. \citet{dunbar2020modelling} analyze compensation in traditional ASR systems based on Hidden Markov Model-Gaussian Mixture Model (HMM-GMM) models. These systems include an \textbf{acoustic model} for mapping acoustic features to phone likelihoods and a \textbf{language model} for modeling the statistical distribution of phone sequences. \citeauthor{dunbar2020modelling} manipulate context-sensitivity in both the acoustic model (monophone vs. triphone) and the language model (flat, unigram, bigram, trigram) and use English and French stimuli containing place or voicing assimilation (e.g., \textit{clea[m] pan}, \textit{ro[p] sale}).
The authors find that human compensation patterns are best predicted by models using contextually-sensitive acoustic models and language models. These models capture allophony and phonotactics, but do not make use of higher-level knowledge of a lexicon or word boundaries. This means that these HMM-GMM models fit better within phonological compensation theories than within lexical compensation theories from the literature. However, the authors solely used stimuli where the assimilation creates a non-word (e.g., \textit{cleam}, \textit{rop}). It could be the case that more challenging cases of assimilation, such as those leading to lexical ambiguity, require higher-level linguistic knowledge for successful compensation.



To our knowledge, assimilation processing remains unexplored in end-to-end neural models for speech recognition. Additionally, no previous work has explored how such models handle lexical ambiguity resulting from assimilation and the potential role of semantic cues in resolving such ambiguities. Despite suggestive correlational evidence that neural ASR models encode several levels of linguistic knowledge, the causal effect of such knowledge on the text transcription task still needs to be demonstrated. 
Our current study hence investigates the causal mechanism behind a previously unexplored phenomenon in neural ASR models, providing novel insights on their internal operations.

\section{Computational Behavioral Experiments}

\paragraph{Experimental Setup} We leverage stimuli from two psycholinguistic studies \cite{darcy2009phonological,gaskell2001lexical}, each targeting the influence of specific linguistic cues on humans' ability to compensate for place assimilation. We specifically focus on \textbf{complete} assimilation. While such complete assimilatory changes may occur less frequently in naturalistic speech than incomplete changes, they are more interesting for our research question: a model must rely on linguistic rather than acoustic cues to compensate for complete changes (see Section \ref{sec:background}).


We feed the stimuli to Wav2Vec2 \cite{wav2vec2}, which is a widely used self-supervised speech model. We use a version of the model that was finetuned for ASR and analyze compensation patterns in its output transcriptions.

\paragraph{Model} Wav2Vec2 consists of a block of seven Convolutional Neural Network (CNN) layers (512 dimensions) and a block of twelve Transformer layers (768 dimensions). The model was pretrained on unlabeled data from the LibriSpeech corpus \cite{librispeech}, which contains 960 hours of English read speech from public domain audio books, sampled at 16 kHz. It takes raw waveforms as input and divides them into frames corresponding to 25 milliseconds of the speech signal, with a 10-millisecond overlap between adjacent frames. Wav2Vec2 is pretrained on predicting latent acoustic information using a contrastive loss function: during pretraining, a random portion of input frames is masked and the model learns to select the correct quantized audio representation from a set of distractors.

We use the \texttt{facebook/wav2vec2-base-960h} implementation from Huggingface \citep{wolf_huggingfaces_2020}, which was finetuned for ASR using 960 hours of labeled data (i.e., transcribed speech) from LibriSpeech. During finetuning, the model learns a mapping between latent frame representations and written characters. Connectionist Temporal Classification (CTC, \citealp{graves2006connectionist}) is used to decode the predicted sequence of characters into well-formed transcriptions---this is necessary because the sequence of characters usually contains duplicates, as most speech sounds span multiple 25-millisecond frames.

\subsection{Experiment 1: Influence of Local Phonological Context}

We first quantify the influence of local phonological context cues on Wav2Vec2's ability to compensate for assimilation. That is, we analyze if the model has learned that assimilation only occurs in specific phonological environments. We use experimental stimuli constructed by \citet{darcy2009phonological}. An important choice in the design of these stimuli is that the assimilated forms are always non-words. This allows us to distinguish between \textbf{phonological compensation} and \textbf{lexical compensation}. With phonological compensation, the model should compensate \textbf{more often} in viable assimilation contexts (e.g., \textit{clea[m] pan}) compared to unviable ones (e.g., \textit{clea[m] spoon}). With lexical compensation, the model can simply recognize that the assimilated form is a non-word (\textit{cleam}) and interpret \textit{clea[m]} as \textit{clean} irrespective of the phonological context that follows. In that case, it should compensate \textbf{equally often} in viable and unviable contexts.

\paragraph{Experimental Design} In the original study, 26 American-English listeners aged 18-53 from the North-East of the United States participated. These participants heard an isolated word without any assimilation (e.g., \textit{clea[n]}), followed by the same word embedded in a carrier sentence. The carrier sentence was constructed according to one of the following three conditions---for clarity, we refer to the final consonant of the target word as `consonant 1', and the initial consonant of the following word as `consonant 2':

\begin{enumerate}
    \item \textbf{Viable assimilation context}: {\it ``Please Sarah, can you hand me a clea[m] pan?''} The articulation place of consonant 1 assimilates to the articulation place of consonant 2.
    \item \textbf{Unviable assimilation context}: {\it ``Please Sarah, can you hand me a clea[m] spoon?''} The articulation place of consonant 1 changes as in the viable condition, but not in line with the articulation place of consonant 2 (this would not occur in natural speech).
    \item \textbf{Control}: {\it ``Please Sarah, can you hand me a clea[n] fork?''} The articulation place of consonant 1 does not change.
\end{enumerate} 

Participants were tasked with identifying whether the isolated target word occurred in the carrier sentence by pressing a "yes" or "no" button as quickly as possible (a button press would then initiate the presentation of a new stimulus; the order of the stimuli was randomized for each participant individually, and they were not allowed to replay stimuli). We let Wav2Vec2 generate transcriptions for each condition instead. To evaluate the model's ability to compensate for place assimilation in each of the three conditions, we measure the \textbf{compensation rate}, which is the percentage of transcriptions for which Wav2Vec2 transcribes the underlying consonant.

The stimuli were recorded by a female native speaker of American English with an accent corresponding to the General American standard. She was instructed to deliberately pronounce the assimilated form (e.g., \textit{clea[m]} in the viable and unviable condition (leading to a complete place assimilation change), and to pronounce the non-assimilated form (e.g., \textit{clea[n]}) in the control condition. The stimuli cover six types of place assimilation, with multiple target words for each type (e.g., \textit{clean}, \textit{own}, \textit{lean}, \textit{thin}, \textit{tan} for /n/ $\rightarrow$ [m]). Each target word was combined with a viable, unviable and control context word, as explained above. Each resulting word pair was then embedded in three different carrier sentences. Thus, for each target word, a total of nine sentences was constructed: 1 target word x 3 context words x 3 carrier sentences. Since there were 16 target words, there were 48 stimuli per condition, and 144 stimuli in total. An overview of assimilation types and number of target words per type is given in Table \ref{tab:assimilation_types} and a full list of stimuli can be found in Appendix B.

\begin{table}[h!]
\small
\centering
\begin{tabular}{llllll}
\toprule
\textbf{Manner} &\textbf{Place} & \textbf{Assimilation} & \textbf{Example} & \#\textbf{Target Words} \\
\midrule
Nasal & Coronal $\rightarrow$ Labial & /n/ $\rightarrow$ [m] & clea[m] pan & 5 \\
Voiced stop &                             & /d/ $\rightarrow$ [b] & ba[b] beer & 3 \\
Voiceless stop &                             & /t/ $\rightarrow$ [p] & fa[p] puppy & 2 \\
Nasal & Coronal $\rightarrow$ Velar  & /n/ $\rightarrow$ [\textipa{N}] & fu[\textipa{N}] game & 3\\
Voiced stop &                            & /d/ $\rightarrow$ [g] & re[g] glasses & 1\\
Voiceless stop &                            & /t/ $\rightarrow$ [k] & grea[k] cruise & 2\\
\midrule
& & & \textbf{Total} & \textbf{16} \\
\end{tabular}
\caption{Overview of place assimilation types in the stimuli by \citet{darcy2009phonological} and the corresponding number of target words per type.}
\label{tab:assimilation_types}
\end{table}

\paragraph{Results}

\begin{figure}
      \centering
      \includegraphics[width=0.75\linewidth]{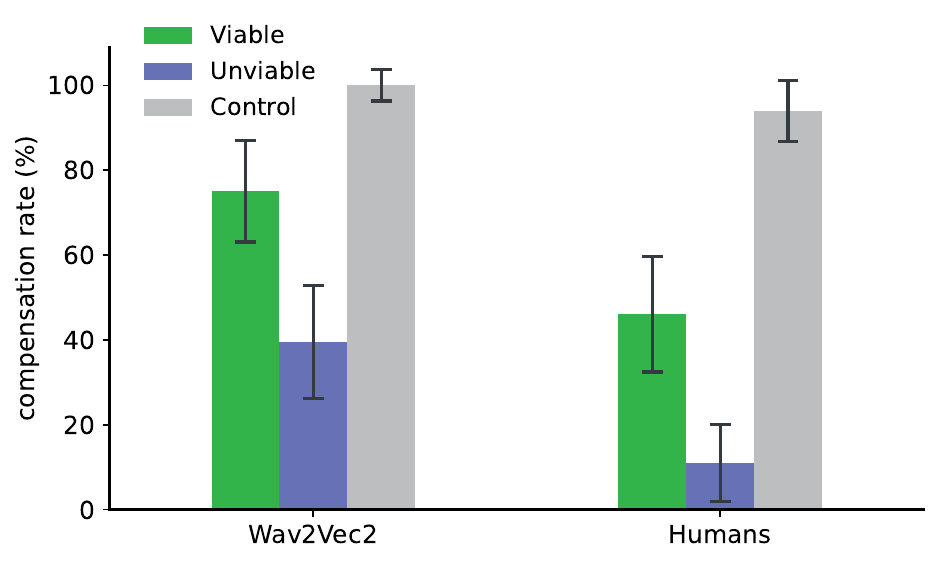}
\caption{Compensation rate (i.e., the proportion of stimuli for which the underlying consonant is transcribed) of Wav2Vec2 and humans in \textbf{viable} and \textbf{unviable} assimilation contexts (e.g., \textit{clea[m] pan} versus \textit{clea[m] spoon}, respectively). In the \textbf{control} condition, the target consonant is not assimilated (e.g., \textit{clea[n] fork}). \textit{N} = 48 for each condition. Error bars denote the 95\% Wilson confidence interval.}
\label{fig:compensation_rate_darcy}
\end{figure}

Figure \ref{fig:compensation_rate_darcy} shows that Wav2Vec2's compensation rate is higher in viable contexts as compared to unviable contexts. This suggests that the local phonological environment in which the assimilation occurs is indeed an important cue for compensation. Despite this, the model still demonstrates a 40 percent compensation rate in unviable assimilation contexts. This suggests that the model also relies on other sources of information to compensate even in situations where assimilation is not expected. These sources of information could be implicitly learned knowledge about valid English words, or knowledge of probable word pairs.



Interestingly, there are cases where the model fails to transcribe the intended word, but still produces another valid lexical candidate rather than the literal spelling of a non-word. For example, for \textit{we[p] pants} (wet pants), it transcribes \textit{we[p]} as \textit{wept} rather than \textit{wep}. Similarly, for \textit{plai[\textipa{N}] condoes} (plain condoes), it transcribes \textit{plai[\textipa{N}]} as \textit{playing} rather than \textit{plaing}. This hints at a reliance on lexical knowledge: the model seems to infer that \textit{wep} and \textit{plaing} are not valid words, and thus transcribes them as the closest lexical candidates (which, according the the model, are \textit{wept} and \textit{playing} rather than \textit{wet} and \textit{plain}). However, it is also possible that the model interprets these forms as outcomes of different phonological processes, such as t-lenition (i.e., weakening of [t] in \textit{wept}, leading to the pronunciation \textit{we[p]}) and vowel reduction (i.e., weakening of [\textipa{I}] in \textit{playing}, leading to the pronunciation \textit{plai[\textipa{N}]}). This would mean that the model is relying on knowledge of various phonological processes beyond place assimilation to process these particular cases.



It could be the case that the compensation rate is higher in the viable condition because the specific bigrams in that condition are seen more frequently during finetuning than those in the unviable condition. To quantify the potential effect of bigram frequency on compensation rate, we compute the Spearman correlation between Wav2Vec2's output probability for the underlying consonant, and the frequency of the target + context word bigrams in the Librispeech finetuning data\footnote{We excluded bigrams containing assimilation from /n/ to [\textipa{N}] from this analysis, because that sound is transcribed as two separate consonants, <n> and <g>. Thus, the correlation analysis is performed on 39 viable stimuli and 39 unviable stimuli (78 in total).}. There is a moderate, positive monotonic correlation between these two variables, $\rho$(76) = .40, \textit{p} < .001. Nevertheless, the model often compensates within bigrams that are never seen during finetuning (see Appendix A for frequencies per bigram). For example, the bigram \textit{mad brother} (pronounced with viable assimilation as \textit{ma[b] brother}) never occurs in the finetuning data, but is still transcribed as \textit{mad brother}. Moreover, the bigram \textit{great match} (pronounced with unviable assimilation as \textit{grea[k] match}) frequently occurs in the finetuning data (43 times), but is still transcribed as \textit{greak match}. This indicates that compensation is not solely driven by bigram frequency.

Finally, we observe that the model compensates generally more often than the human participants in \citeauthor{darcy2009phonological}'s study. This could be a task-specific effect---humans had to directly compare non-assimilated forms in isolation with assimilated forms in sentential context, whereas the model had to do a transcription task. Therefore, humans might have been more sensitive to subtle acoustic differences between the isolated and embedded word. Another explanation is that humans relied less on lexical knowledge than the model seems to be doing. We further dissect the role of phonological and lexical cues in the following section.

\subsection{Experiment 2: Isolating Phonological and Lexical Compensation}
\label{exp2}

The results from Experiment 1 indicate that compensation for place assimilation is sensitive to local phonological cues, but might additionally be guided by other sources of information, including lexical knowledge and bigram frequency. Knowledge of valid word candidates and probable word pairs might be used to "override" non-words with valid words. To isolate phonological compensation from lexical override of non-words, we conduct a follow-up experiment using stimuli by \citet{gaskell2001lexical}. These stimuli again contrast viable and unviable assimilation contexts, but with the assimilated form being a potential lexical candidate (not a non-word, as in our previous experiment).

\paragraph{Experimental Design} The original study used a cross-modal priming paradigm in which 83 British-English listeners aged 18-45 participated. They heard sentences in the following conditions:

\begin{enumerate}
    \item \textbf{Viable assimilation context}: \textit{I think a quick ru[m] picks you up.} Here, the word \textit{ru[m]} is potentially ambiguous --- it can either be the standard pronunciation of \textit{rum}, or the assimilated form of \textit{run}, with the final /n/ assimilated to the following labial [p] in \textit{picks}.
    \item \textbf{Unviable assimilation context}: \textit{I think a quick ru[m] does you good.} Here, the word \textit{ru[m]} is not ambiguous --- it can only be the standard pronunciation of \textit{rum}, since the [d] in \textit{does} does not trigger assimilation from \textit{run} to \textit{ru[m]}.
\end{enumerate}

At the offset of the spoken prime word (e.g., \textit{ru[m]}), a visual word appeared on a screen in front of the participant for 200 ms, for which they had to make a lexical decision (they had to press a "yes" or "no" button indicating whether the visual word was a valid word or not). This visual word would either match the surface form, e.g., \textit{rum}, or the underlying form, e.g., \textit{run}. Response times for these lexical decisions were analyzed to determine whether the participants compensated for the assimilation or not---quicker response times for the underlying form would suggest compensation. We again measure the compensation rate in Wav2Vec2's output transcriptions, as in Experiment 1.


Since the original stimuli by \citet{gaskell2001lexical} were not available, we asked a female native speaker of American English to record them.\footnote{The fact that the stimuli of Experiment 1 and 2 were recorded by different speakers in different labs may introduce confounds. To minimize this possibility, we aimed to make our stimuli as comparable as possible to those from \citeauthor{darcy2009phonological}. This involved selecting a speaker of a similar American English accent, despite the original stimuli from \citeauthor{gaskell2001lexical} being designed for British English. As such, we excluded stimuli that did not work as a minimal pair in American English, such as \textit{balm/barn}. Additionally, we repeated Experiment 1, 2 and 3 with stimuli pronounced by a different native speaker of American English (male). The results are highly similar (see Appendix A), which indicates that the observed compensation behavior in Wav2Vec2 is robust across speakers.} The speaker was instructed to 1) deliberately pronounce the assimilated form (e.g., \textit{rum}), 2) to connect it with the following word (e.g., \textit{picks}) and 3) to not put any unnatural emphasis on the crucial word pair. The stimuli were presented one by one on slides and the speaker recorded them in a quiet booth using a Blue Yeti X microphone. Stimuli were recorded and saved at 44100 Hz, and downsampled to 16000 Hz by HuggingFace's preprocessor before being fed to the model.

We recorded each stimulus item by having the speaker read the sentence containing assimilation preceded by a biasing context sentence (see \autoref{exp3}). The recording was split per item and annotated in Praat \citep{praat}; TextGrid annotations indicated the intervals containing each sentence. The final stimulus audio files were created with a custom Python script using the TextGridTools \citep{buschmeier_textgridtools_2013} and Parselmouth packages \citep{jadoul_introducing_2018}. We ensured that the audio for each stimulus item was equal in length by padding the sentence audio with background silence (recorded at the same location). As such, each stimulus audio fed as input to the model was exactly 8 seconds long. 
For Experiment 2, we added 150 ms of silence after the viable and unviable target sentences and filled the remaining duration with silence preceding the start of the sentence.

\paragraph{Results} 

\begin{figure}
      \centering
      \includegraphics[width=0.75\linewidth]{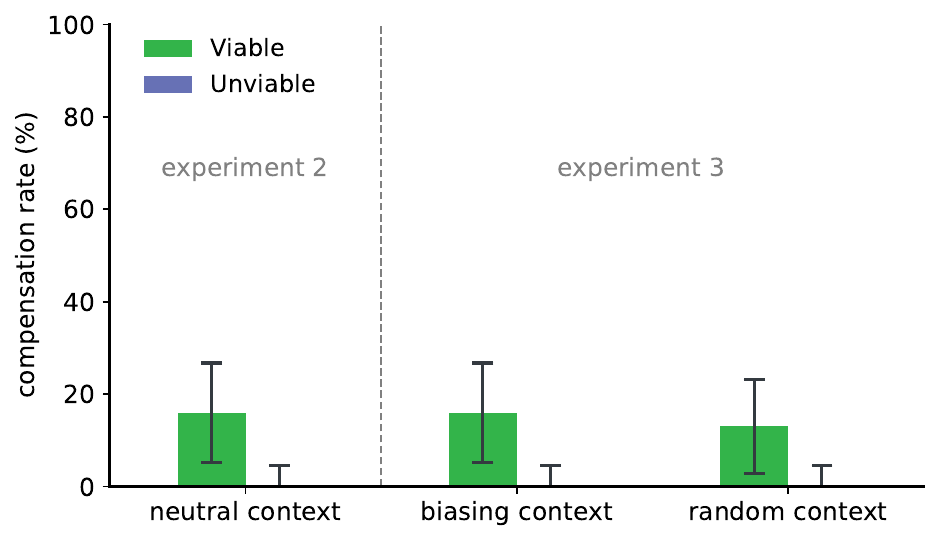}
\caption{Compensation rate (i.e., the proportion of stimuli for which the underlying consonant is transcribed) of Wav2Vec2 in \textbf{viable} and \textbf{unviable} assimilation contexts (e.g., \textit{ru[m] picks} versus \textit{ru[m] does}, respectively), with different types of preceding sentential context (\textbf{neutral context}, \textbf{biasing context}, and \textbf{random context}). \textit{N} = 38 for each condition. Error bars denote the 95\% Wilson confidence interval.}
\label{fig:compensation_rate_gaskell}
\end{figure}


The results are presented in Figure \ref{fig:compensation_rate_gaskell} (under "experiment 2"). We observe a similar pattern as for Experiment 1: the model compensates more often in viable assimilation contexts as compared to unviable ones. This again suggests that Wav2Vec2's compensation mechanism is sensitive to phonological context cues. Nevertheless, the overall compensation rates of Experiment 2 are much lower than those measured in Experiment 1. This provides further evidence that the results of Experiment 1 were additionally guided by the lexical override of non-words---a process that cannot happen if the assimilated form is a valid lexical candidate. In the latter case, the model can theoretically compensate based on phonological cues alone, but mostly adopts the surface interpretation in practice (which explains the relatively small difference between the viable and unviable condition compared to the difference observed for Experiment 1).



\subsection{Experiment 3: Influence of Biasing Sentential Context}
\label{exp3}

Our results thus far indicate that Wav2Vec2 uses the local phonological environment of the assimilated sound to infer the underlying phoneme. When the assimilated form is a non-word (as in Experiment 1), there appears to be a strong additional reliance on lexical knowledge in the compensation process. When the assimilated form is a competing lexical candidate (as in Experiment 2), Wav2Vec2 mostly does not consider that assimilation is an option and just transcribes the surface form (given that the underlying and surface form are equally likely in the semantic context). But what if the semantic context makes the underlying form more likely?



In our final behavioral experiment, we analyze if biasing sentential context increases Wav2Vec2's compensation rate. If the model indeed incorporates semantic context, it should compensate more often when there is biasing sentential context compared to neutral or random (non-biasing) context. If Wav2Vec2 simply adopts the surface interpretation without considering semantic context cues, the model should compensate equally often in all sentential contexts.

\paragraph{Experimental Design} We constructed the following three experimental conditions using the stimuli by \citet{gaskell2001lexical}:

\begin{enumerate}
    \item Viable assimilation in a \textbf{neutral context} (with preceding silence): \textit{[silence] I think a quick ru[m] picks you up.}
    \item Viable assimilation with preceding \textbf{biasing context}: \textit{It’s best to start the day with a burst of activity. I think a quick ru[m] picks you up.}
    \item Viable assimilation with preceding \textbf{random context}: \textit{We were impressed by her stylish delivery and intonation. I think a quick ru[m] picks you up.}
\end{enumerate}

\noindent We also constructed these conditions for the unviable stimuli (e.g., \textit{It's best to start the day with a burst of activity. I think a quick ru[m] does you good}). The original study did not include the random context condition, but we included this as an additional check---does biasing context influence compensation patterns in any way, and if so, does biasing context have a different influence than \textit{any} random context?

The stimuli for Experiment 3 were recorded by the same speaker and constructed using the same script described for Experiment 2 (see Section \ref{exp2}). We again padded the sentence recordings with background silence to ensure each stimulus item was 8 seconds long: for the stimuli with a preceding (biasing or random) context sentence, we inserted 250 ms of silence between sentences, 150 ms before and after the sentence pair, and filled the remaining duration with silence preceding the start of the first sentence. We used the same context sentence recordings in constructing stimuli for both the viable and unviable conditions. For the random context condition, target sentences were randomly paired with a context sentence from another stimulus item, ensuring that each context sentence was only used once across all random sentence pairs.

\paragraph{Results} 

Figure \ref{fig:compensation_rate_gaskell} shows that Wav2Vec2 compensates equally often in all three sentential contexts. This indicates that acoustic information has priority, even when semantic context biases away from the lexical candidate matching the speech waveform.

It could be the case that, even if the surface form ultimately "wins", the model still considers the underlying form. We checked for the potential activation of both lexical candidates in Wav2Vec2 by measuring if the probability for the underlying phoneme increases in the presence of biasing context. We found no difference between this probability in the presence of biasing context (\textit{M} = 0.14, \textit{SD} = 0.30) and neutral context (\textit{M} = 0.17, \textit{SD} = 0.30). This provides further evidence that the model does not integrate long-distance semantic context when processing assimilated inputs.

\paragraph{Overall comparison with human behavior}

\begin{figure}
      \centering
      \includegraphics[width=0.99\linewidth]{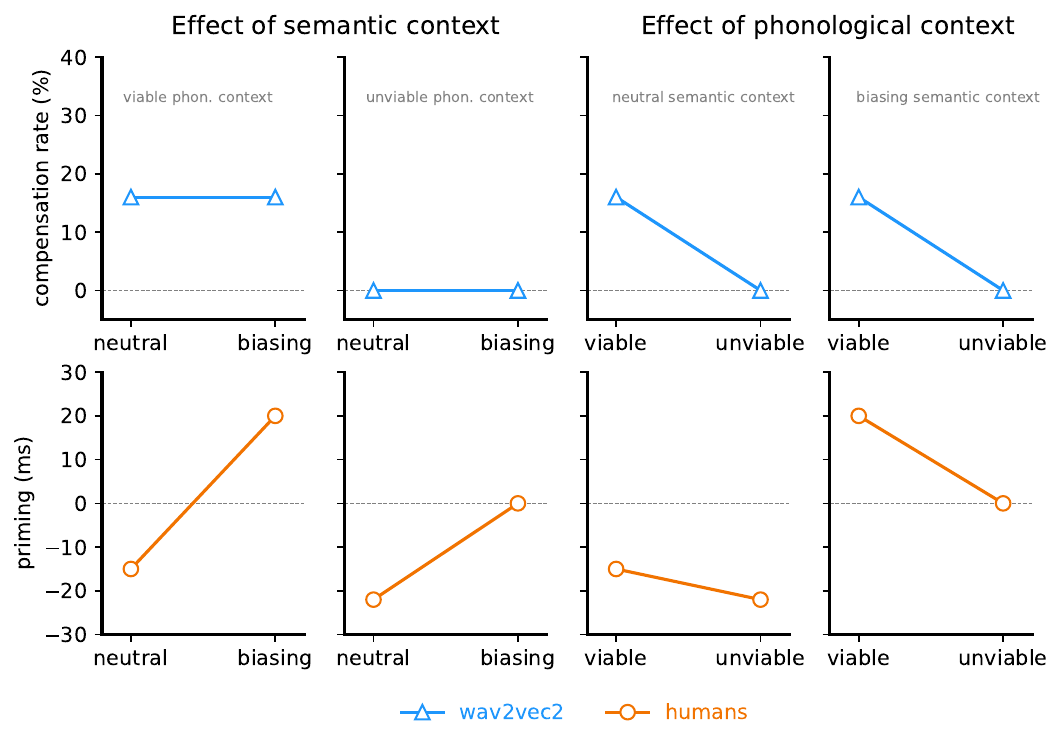}
\caption{Comparison between compensation behavior of Wav2Vec2 and human participants from \citet{gaskell2001lexical}. The left side of the figure shows the effect of \textbf{semantic context} (neutral vs. biasing) on compensation behavior; the right side of the figure shows the effect of \textbf{phonological context} (viable vs. unviable).}
\label{fig:gaskell-hum-vs-w2v}
\end{figure}

Wav2Vec2's transcription task differs from the lexical decision task that humans performed in the original study by \citeauthor{gaskell2001lexical}, complicating a direct comparison of results across the two methodologies. Nevertheless, there appear to be interesting parallels and differences between human and model behavior. In Figure \ref{fig:gaskell-hum-vs-w2v}, we visualize Wav2Vec2's behavior by its \textbf{compensation rate} in different semantic and phonological contexts (neutral vs. biasing, and viable vs. unviable), and we visualize human participants' behavior by their \textbf{priming effects} in these contexts.

We observe a clear difference in the effect of semantic context on model and human compensation patterns. Humans clearly compensate more in the presence of biasing context than in neutral semantic context. The model, however, is not affected by such semantic context and compensates equally often in neutral and biasing environments. Phonological context, on the other hand, has a similar effect on model and human behavior: both compensate more if the assimilation is viable in the phonological context.

Overall, we can characterize the observed behavior as follows: when the surface and underlying interpretation are equally likely in the semantic context, neither humans nor the model will "go through the effort" of considering the underlying lexical candidate, even if assimilation could theoretically make that candidate possible (note the negative priming effects for humans and the low compensation rates for the model in neutral semantic context). When semantic context biases away from the surface interpretation, humans will consider both lexical candidates, but they will only choose the underlying candidate if the assimilation is licensed in the phonological context (note the priming effect of 20 ms in the viable condition and the priming effect of 0 ms in the unviable condition). Wav2Vec2, on the other hand, will not consider both candidates and will stick to the interpretation that most closely follows the speech waveform.



\subsection{Discussion}

Taken together, our results offer some implications for the theories described in Section \ref{sec:theories} in the context of a neural ASR system. Experiment 1 and 2 showed that Wav2Vec2 is sensitive to phonological context, as compensation rates were higher in phonologically viable contexts than phonologically unviable contexts. This pattern aligns with human behavior and corroborates the idea that compensation is not purely driven by lexical knowledge. Nevertheless, we find a clear asymmetry in compensation patterns for non-words and real words: the model compensates much more frequently if the assimilation transforms the lexical item into a non-word (e.g., \textit{clea[m]}) instead of an alternative lexical candidate (e.g., \textit{ru[m]}). A purely phonological compensation mechanism cannot explain this asymmetry.


Experiment 3 showed that the introduction of semantic context did not increase compensation rates of the model, which is not entirely in line with earlier findings of human participants. This might be caused by
Wav2Vec2's training objective, which centers around frame-level predictions (i.e., reconstructing masked frames in the pretrained version, and predicting frame-level characters in the finetuned version). To make such predictions, there is often no need to integrate long-distance contextual information from previous sentences. For this study, we purposefully chose a model without an explicit language modelling component because we aimed to elicit implicitly learned linguistic knowledge. Nevertheless, it would be of interest for future work to conduct comparisons with models featuring an explicit language modeling objective, since these models may potentially be better at capturing longer-distance semantic relations between tokens.\footnote{We conducted an initial exploration regarding the role of explicit language modelling using the encoder-decoder model \texttt{Whisper} \cite{radford2023robust}, in which the decoder performs next-token prediction. We found that this model generally compensates more often than Wav2Vec2, but is still not affected by the preceding context.}




\section{Interpretability Experiments}

The behavioral experiments described in the previous sections offer valuable insights into the types of cues that neural ASR systems use in compensating for place assimilation. However, given the inherent challenge of tracing the input-output information flow in black-box neural models, we cannot determine the precise implementation of the compensation mechanism from behavioral experiments alone. 
What we can do, though, is analyze the hidden components of our computational model, using techniques from interpretability research \cite{lyu2024towards}.

In the remainder of this study, our focus narrows to a more thorough understanding of the compensation process in cases such as \textit{clea[m] pan}. Our results suggest that both phonological cues and lexical knowledge may contribute to compensation in such cases. We conduct a series of interpretability experiments to better understand this. Firstly, we \textit{localize} the layers where compensation seems to occur. Then, we take first steps at \textit{mechanistically} understanding how compensation is achieved. This involves analyzing which context cues causally contribute to the final transcription and identifying which subregions of the network seem to propagate these cues to the assimilated consonant.

\subsection{Localizing Compensation}

\paragraph{Methodology} We extract layer-wise frame representations from Wav2Vec2 at the position of the assimilated consonant. We use the model's output transcription to find that position. For example, we extract the frame representation for which the character <n> was predicted if the model transcribes \textit{clean}, and the frame representation for which the character <m> was predicted if the model transcribes \textit{cleam}\footnote{If the model predicted the character for more than one frame, we used the first of those frames as our probing input. For cases where <ng> was predicted (e.g., fu[\textipa{N}] night $\rightarrow$ fu<ng> night), we also used the first occurrence of <n> as our probing input.}. We then feed this representation into a probing classifier that was trained to distinguish between the (underlying) phonemes /n/ and /m/ using data from an external corpus (TIMIT, see training details below). We analyze how the probability for the underlying consonant (/n/ in this case) changes across layers. This process is repeated for all assimilation types listed in Table \ref{tab:assimilation_types}, which means that we train an individual binary probing classifier for each assimilation type.

We analyze probability patterns for the viable and unviable stimuli separately and divide them into two categories: \textit{compensation} and \textit{no compensation}. This refers to whether the model compensated for the place assimilation in its final transcription. We also perform the analysis on the control stimuli.
 
\paragraph{Probing Classifiers} 

We train Logistic Regression classifiers to decode phoneme identity from frame representations extracted from each Transformer layer. We use the TIMIT Acoustic-phonological Continuous Speech Corpus \cite{AB2/SWVENO_1993}, which contains sentence recordings of 630 speakers of eight major American-English dialects. Each speaker read aloud the same ten sentences, which were designed to elicit a wide variety of (dialect-specific) speech sounds. The corpus includes time-aligned transcriptions of phonemes and words, as well as the raw waveform for each spoken sentence, sampled at a rate of 16 kHz.

\begin{figure}
      \centering
    \includegraphics[width=0.75\linewidth]{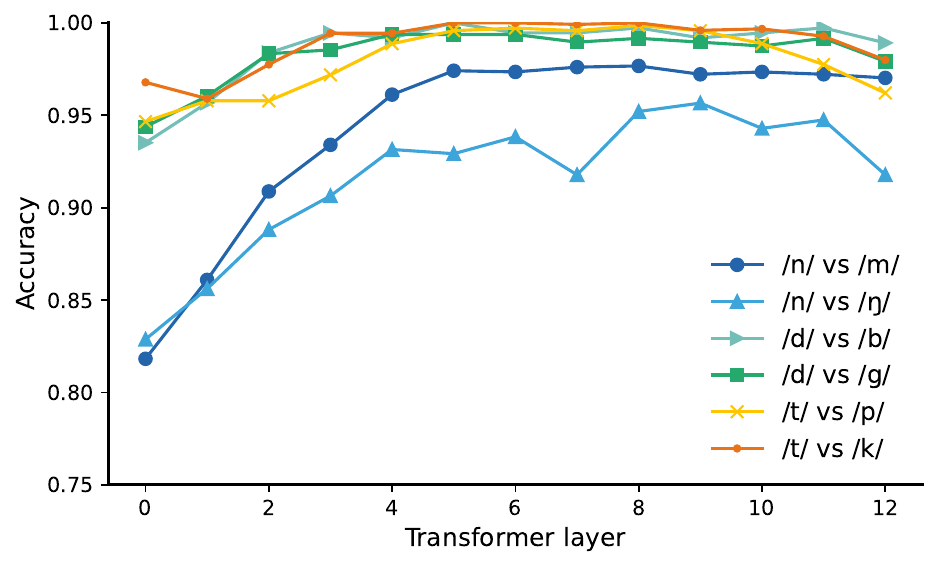}
\caption{Accuracy of binary probing classifiers, trained and evaluated on frame-level representations from individual Wav2Vec2 layers (extracted using the TIMIT corpus). Each classifier has to distinguish between two candidate phoneme labels (indicated in the legend).}
\label{fig:probing-accs}
\end{figure}

To generate the training and evaluation data, we pass 1000 utterances from the TIMIT train split and 200 utterances from the TIMIT test split through Wav2Vec2 and label the corresponding frame representations using the phoneme timestamps in TIMIT. We do this in a maximalist fashion: all frames in which (part of) a phoneme occurs are labeled with that phoneme. Each probing classifier is trained on representations from a single Wav2Vec2 layer, and only has to distinguish between two phoneme labels (i.e., the underlying and surface consonant involved in the assimilation process). To make sure that the train and test data for each probing classifier is balanced, we downsample the majority phoneme class to match the number of samples in the minority phoneme class. All probing models reach high accuracy, especially when trained on middle to final layers (see Figure \ref{fig:probing-accs}).

\begin{table}[h!]
\small
\centering
\begin{tabular}{cccc}
\toprule
\textbf{Phoneme Contrast} & \#\textbf{Train Frames per Class} & \#\textbf{Test Frames per Class} \\
\midrule
/n/ vs /m/ & 3325 & 773 \\
/n/ vs /\textipa{N}/ & 1157 & 197 \\
/d/ vs /b/ & 865 & 204 \\
/d/ vs /g/ & 1116 & 220 \\
/t/ vs /p/ & 1688 & 356 \\
/t/ vs /k/ & 3165 & 622 \\
\end{tabular}
\caption{Number of train and test frames for each binary probing classifier after downsampling the majority phoneme class to the minority phoneme class. Frames are obtained by running utterances from the TIMIT corpus through Wav2Vec2.}
\label{tab:probing-classifiers}
\end{table}

\paragraph{Results} 

\begin{figure*}
      \centering
      \includegraphics[width=1\linewidth]{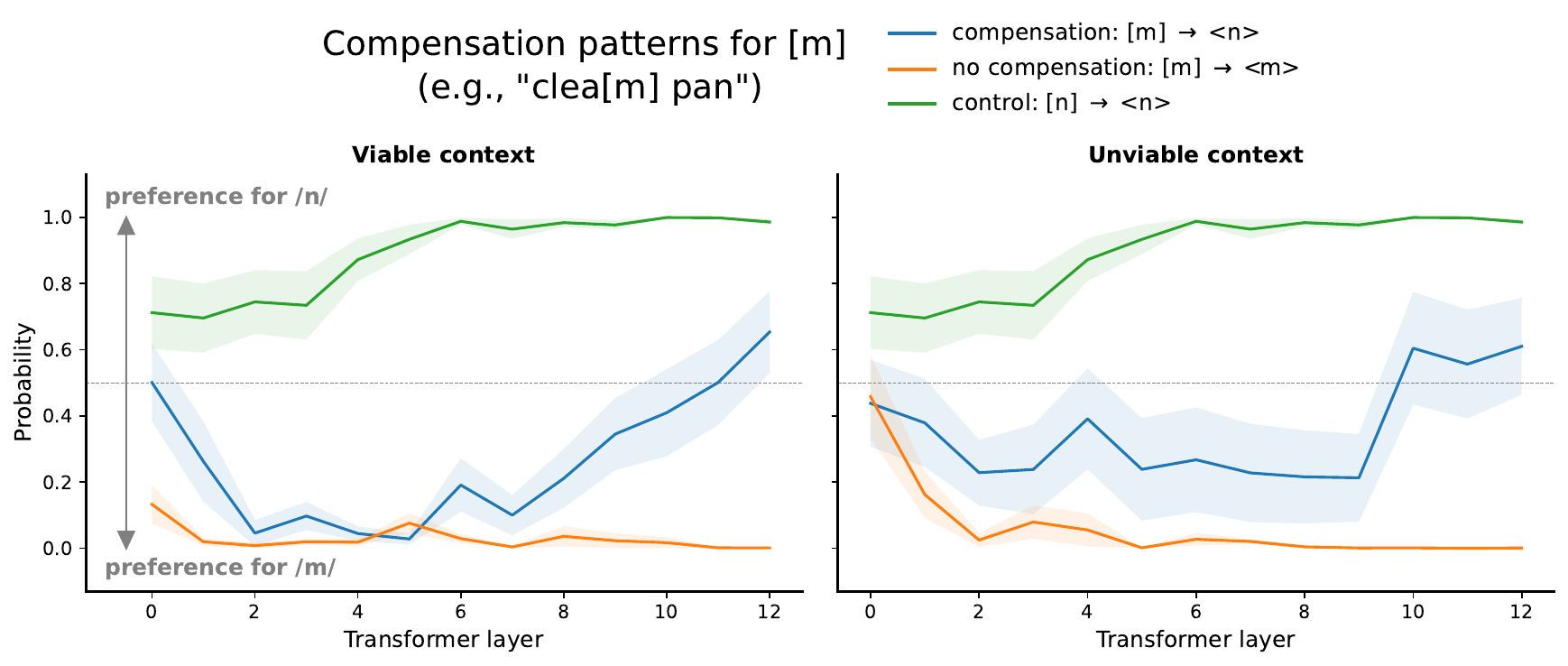}
      \includegraphics[width=1\linewidth]{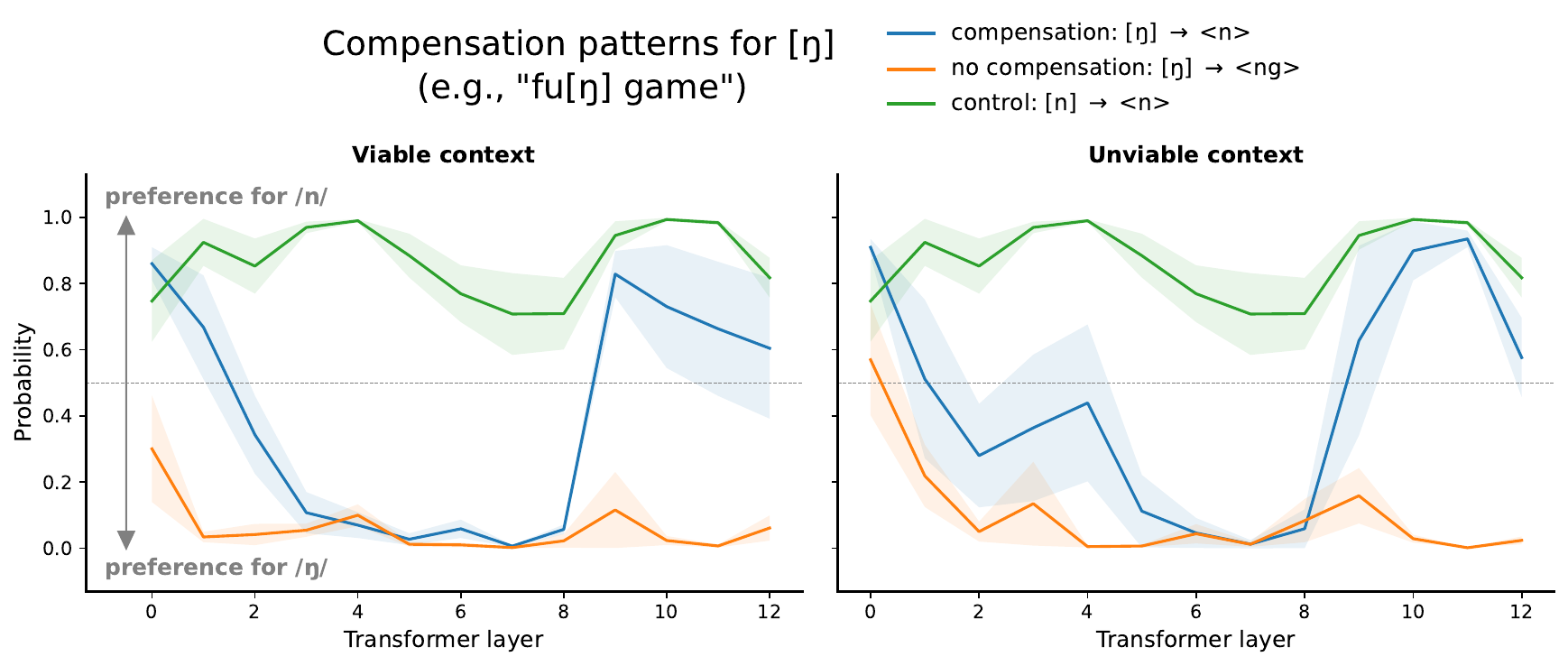}
\caption{Layerwise preference of binary linear probing classifiers for the underlying consonant /n/ or the surface consonant (top: /m/, bottom: /\textipa{N}/) given Wav2Vec2 representations at the position of the assimilated consonant. The three line colors indicate whether the model compensated for the assimilation in its final transcription. Error bars denote the standard error of the mean.}
\label{fig:probing-nasals}
\end{figure*}

Figure \ref{fig:probing-nasals} shows the layer-wise probability patterns for the nasals. The results of the stops are comparable and are shown in Appendix C. We observe distinct layer-wise patterns depending on the compensation behavior in the model's final transcription. When the model fails to compensate in its transcription (orange), it seems to interpret the assimilated sound as the surface form [m] or [\textipa{N}] across all layers. When the model succeeds in compensating (blue), however, the model seems to first interpret the assimilated sound as the surface form [m] or [\textipa{N}] , but later shifts its interpretation to the underlying form /n/. In the control condition (green), when the input consonant is not assimilated and thus pronounced as [n], the model interprets the sound as /n/ across all layers.\footnote{We repeated the probing experiments with the pretrained version of Wav2Vec2, \texttt{facebook/wav2vec2-base}, and observed similar probability patterns as for the finetuned model. The only exception were the probabilities in the final layers, which showed a converging instead of diverging pattern across conditions. This can potentially be explained by the different model objectives: While the finetuned model has to actively distinguish between different characters, the pretrained model has an autoencoder-style objective of closely reconstructing masked input frames \cite{pasad2021layer}.}

The blue pattern suggests a distinct functionality of the model's layers. Early to middle layers seem to be identifying the surface phones that are pronounced (this identification is not completely accurate yet in the first two layers, in line with the layer-wise accuracies depicted in Figure \ref{fig:probing-accs} and corresponding to the findings of \citet{pasad2021layer}, who show that Wav2Vec2's middle layers correlate strongly with phone identity). Later layers seem to have incorporated contextual information, so that the model can infer the phonemes underlying the surface sounds. In the following section, we try to trace how the model collects relevant context cues to achieve the shift from surface to underlying interpretation in its final layers.



\subsection{Tracing the Compensation Mechanism using Causal Interventions}


In our final set of experiments, we explore whether 
the behavior \textit{compensation for place assimilation} can be traced back to specific subregions of Wav2Vec2. We analyze the causal contribution of specific context cues to the final character prediction at the position of the assimilated consonant. Moreover, we analyze which model components in which layers are involved in propagating these cues to the target position.

\paragraph{Causal Interchange Interventions}

We perform \textbf{causal interchange interventions} within the 12 Transformer layers of Wav2Vec2, each containing 12 attention heads and a Multi-Layer Perceptron (MLP). These model components can be thought of as 'information movers', which read from and write to the \textit{residual stream} of individual frames \cite{wang2022interpretability}. In text-based language models, causal interventions have been used to identify subgraphs of model components (referred to as \textit{circuits}) responsible for \textit{greater-than} reasoning \cite{hanna2024does}, pronoun resolution in Winograd sentences \cite{yamakoshi-etal-2023-causal}, and gender bias \cite{vig2020investigating,chintam2023identifying}. Here, we apply causal interventions to identify similar potential subgraphs underlying Wav2Vec2's compensation behavior.


\paragraph{Methodology} We start by identifying a pair of viable and unviable stimuli from Experiment 1 for which the model showed the desired output behavior; that is, it should transcribe the underlying consonant in the viable condition, and the surface consonant in the unviable condition. An example of such a pair is \textit{thi[m] packet} and \textit{thi[m] leaflet}, for which the model transcribed \textit{thi\textbf{n} packet} and \textit{thi\textbf{m} leaflet}. 

\begin{figure}
      \centering
      \includegraphics[width=1\linewidth]{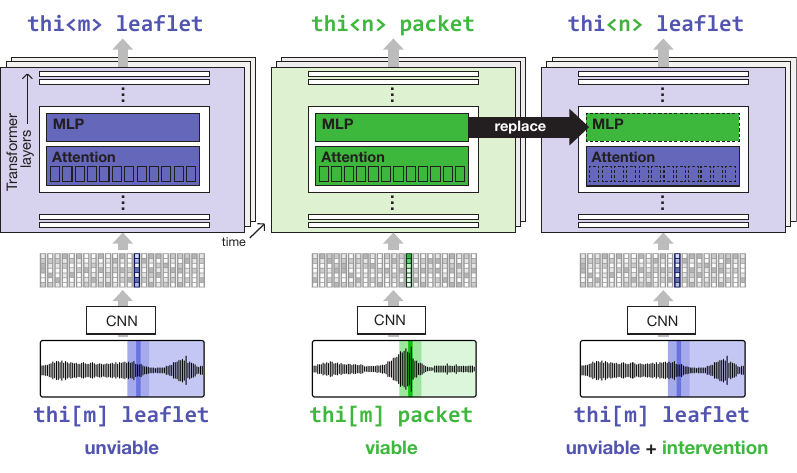}
\caption{Causal interventions were performed at the critical frame, critical phone, and whole word level (indicated in dark-to-light shading in the audio signal), and for several different components within individual Transformer layers (outputs of attention heads and MLPs, indicated in dashed frames). Here we visualize the replacement of an MLP output of the unviable run with an MLP output from the viable run, for the critical frame of the context word \textit{leaflet/packet} only.}
\label{fig:causal-interventions}
\end{figure}

A schematic overview of the intervention procedure is depicted in Figure \ref{fig:causal-interventions}. We run the model on both inputs, which we will henceforth refer to as the \textbf{viable run} (shown in the middle in green) and the \textbf{unviable run} (shown on the left in blue). During both runs, we extract the output of all individual attention heads and MLPs across all frames. Next, we repeat the unviable run, but this time, we replace the output of a specific head or MLP with the output that was generated by that component during the viable run (shown on the right in blue and green). After each interchange intervention, we analyze if the probability of the critical frame changed (i.e., the frame for which <m> was predicted). We are particularly interested in interventions that are able to flip the prediction from <m> to <n>, as these should reveal the crucial model components responsible in compensating for assimilated phonemes.

The importance of individual attention heads can be broken down into the importance of its subcomponents (keys, queries, values). The \textbf{full head output vector} is essentially a summary of relevant context cues from various positions. By measuring the effect of replacing this output vector, we assess how much relevant context information is introduced into the residual stream by that head at that layer as a whole. Then, to get a more fine-grained picture of the specific contextual cues that are propagated by that head, we intervene on the \textbf{value vector} of that head at various frame positions.

\begin{figure}
      \centering
      \includegraphics[width=1\linewidth]{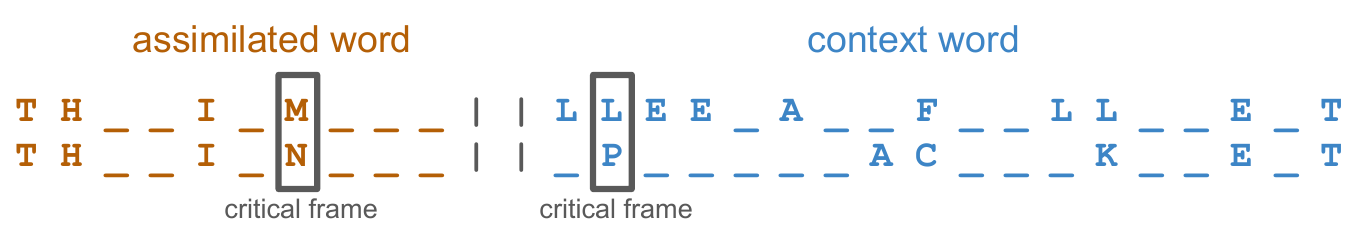}
\caption{Predicted characters for the unviable input \textit{thi[m] leaflet} and the viable input \textit{thi[m] packet}. The critical frames for which we replace head or MLP activations are marked.}
\label{fig:transcripton-example}
\end{figure}

To quantify the importance of different types of context cues (phonological and/or lexical), we perform the interventions at six different positions. To illustrate these positions, we showcase Wav2Vec2's transcription for the unviable input \textit{thi[m] leaflet} and the viable input \textit{thi[m] packet} in Figure \ref{fig:transcripton-example}. For both the assimilated word and the context word, we replace the activations at the position of the \textbf{critical frame}, \textbf{critical phone} (the critical frame plus the three frames preceding and following it), and the \textbf{whole word}.

\paragraph{Results}

\begin{figure}
      \centering
      \includegraphics[width=1\linewidth]{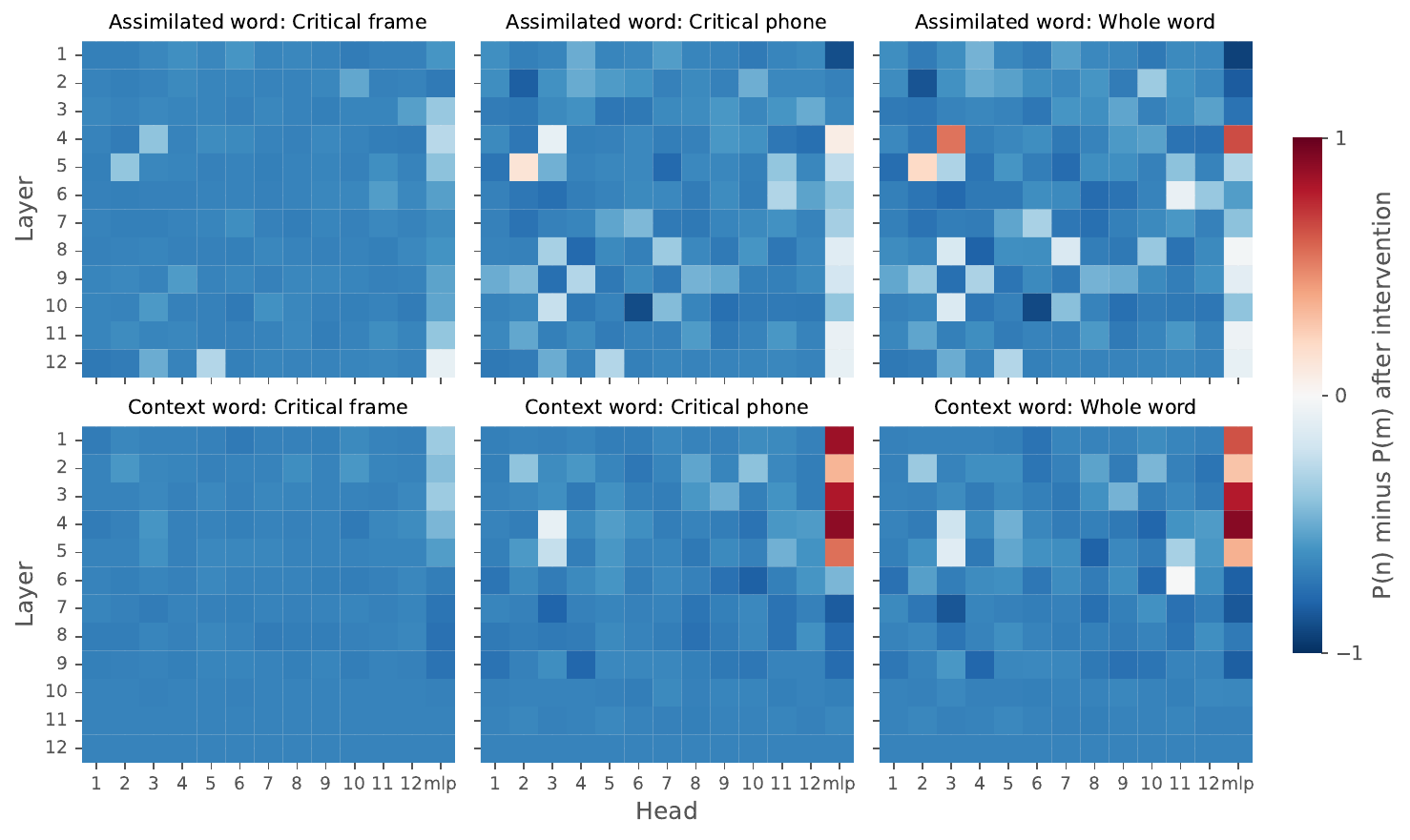}
\caption{Probability difference between the underlying consonant label <n> and the surface consonant label <m> after replacing individual \textbf{head or MLP outputs} at different positions (titles indicate the intervention position). We swap in the activations corresponding to the viable utterance \textit{thi[m] packet} while running the finetuned Wav2Vec2 model on the unviable utterance \textit{thi[m] leaflet}. If the probability difference is positive (red), the model's transcription is flipped from \textit{thim} to \textit{thin}.}
\label{fig:head-mlp-interventions}
\end{figure}

In Figure \ref{fig:head-mlp-interventions}, we showcase the impact of replacing individual head and MLP outputs from the unviable run (\textit{thi[m] leaflet}) with those from the viable run (\textit{thi[m] packet}). The positions at which we replace activations are denoted by the titles of each heatmap.

We observe a drastic increase in the probability for the underlying consonant <n> when intervening on early MLP outputs (layers 1-5) at the context word position \textit{leaflet/packet}. This effect is absent when intervening solely on the critical context frame (i.e., the single frame for which <l> or <p> was transcribed, as seen in Figure \ref{fig:transcripton-example}). Instead, the effect becomes evident when intervening on all frames spanning the critical context phone [l]/[p]. Notably, the effect does not become stronger when intervening on the entire context word \textit{leaflet/packet}. This suggests that the context \textbf{phone} serves as the "determining" cue that leads the model to transcribe <n> rather than <m>. Early MLPs apparently write this cue to the residual stream of the critical target frame (i.e., the frame for which <m> or <n> is transcribed, as seen in Figure \ref{fig:transcripton-example}).

This observation may also clarify why intervening on later MLPs (in layers 6-12) is effective only within the assimilated word \textit{thi[m]}, not within the context word \textit{leaflet/packet}. By that point, the relevant context cue has already been embedded in the residual stream of assimilated frames. Consequently, in later layers, the probability can only be shifted from <m> to <n> by directly modifying activations within the assimilated word \textit{thi[m]}. This would also explain the observed shift from surface to underlying interpretation in the model's final layers, as classified by probing models (see Figure \ref{fig:probing-nasals}).

Shifting our focus to the attention heads in Figure \ref{fig:head-mlp-interventions}, we notice that replacing the outputs of two specific heads (i.e., head 3 in layer 4 and head 2 in layer 5) can switch the prediction from <m> to <n>. This effect is observable only when swapping those outputs at the assimilated word position \textit{thi[m]}, not the context word position \textit{leaflet/packet}---which seems to be the opposite behavior of the early MLPs.

\begin{figure}
      \centering
      \includegraphics[width=1\linewidth]{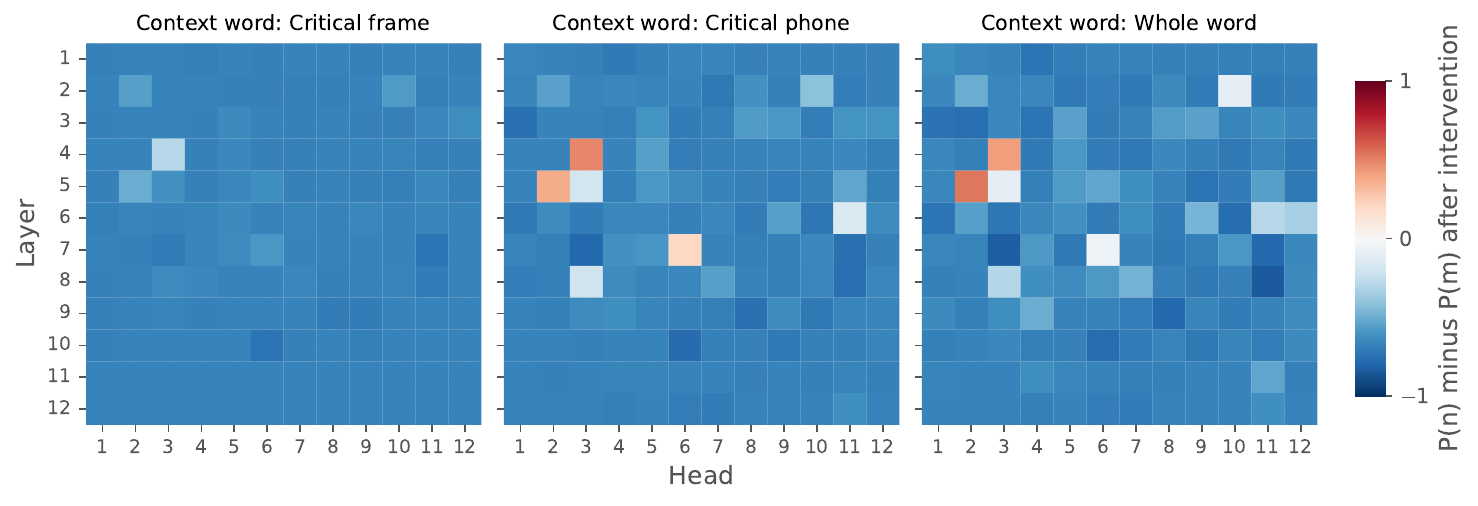}
\caption{Probability difference between the underlying consonant label <n> and the surface consonant label <m> after swapping the \textbf{value vectors} of individual heads at different positions (titles indicate the intervention position). We swap in the activations corresponding to the viable utterance \textit{thi[m] packet} while running the finetuned Wav2Vec2 model on the unviable utterance \textit{thi[m] leaflet}. If the probability difference is positive (red), the model's transcription is flipped from \textit{thim} to \textit{thin}.}
\label{fig:val-vector-interventions}
\end{figure}

Recall that an attention head's output essentially serves as a summary of relevant context cues from various positions. It is plausible that head 3 in layer 4 and head 2 in layer 5 have a large effect on the probability because they predominantly gather information from the critical context phone [l]/[p] in \textit{leaflet/packet}. To validate this, we intervene on the \textbf{value vectors} of relevant context frames, as shown in Figure \ref{fig:val-vector-interventions}. Notably, these value vector interventions yield a similar effect as head output interventions at the assimilated word. This suggests that head 3 in layer 4 and head 2 in layer 5 write information about the critical context phone into the residual stream of the assimilated word \textit{thi[m]}.

Similar to our findings for the MLPs, intervening on the value vectors of the critical context phone [l]/[p] has a similar effect as intervening on the value vectors of the entire context word \textit{leaflet/packet} (compare the middle and right heatmap in Figure \ref{fig:val-vector-interventions}). This implies that phonological cues trigger compensation for assimilation in this specific example.

We conducted the same analysis for several other examples and observed that the pattern of the MLPs remain consistent: intervening on the critical context phone is effective for early MLPs, while intervening on the assimilated consonant is effective for later MLPs. The patterns observed in the attention heads, however, exhibit more variability across examples. Further research and, ideally, the development of an automatic algorithm for determining important model components are required to ascertain if \textit{compensation for place assimilation} can be traced back to a unified circuit. Additionally, quantifying the interactions between components is desirable, as our current approach only evaluates the importance of model components individually.

\section{Conclusion and Discussion}

In this study, we have conducted a systematic analysis of the mechanism employed by Wav2Vec2 to address place assimilation. Through computational behavioral experiments, we have identified the specific linguistic environments in which the model compensates, and we observed interesting parallels and differences with human compensation patterns. We found that, like humans, the model is sensitive to the local phonological context in which the assimilation occurs, and rarely compensates when the phonological context does not license assimilation. The model's compensation patterns seem to be further guided by lexical knowledge, but the model does not integrate semantic context. Through probing experiments and causal interventions, we pinpointed the model components in which compensation occurs, and demonstrated the causal role of phonological cues in the compensation process.

We observed that context cues that are as minimal as a single phone can have a drastic effect on the output of a neural ASR model. This sparks curiosity about the role of such minimal cues in human assimilation processing. \citet{mitterer2003coping} studied the timecourse of compensation for assimilation by exposing Dutch participants to compound words containing viable assimilation (e.g., \textit{tui[m]bank}, "garde[m] bench") and unviable assimilation (e.g., \textit{tui[m]stoel}, "garde[m] chair"). Analyzing participants' EEG signals, they found a mismatch negativity at the onset of the context word, indicating that participants perceived [m] differently as a result of the following phone, which either licensed assimilation ([b]) or not ([s]). It would also be interesting to present human subjects with increasing amounts of material from the context word (i.e., \textit{garde[m] b...}, \textit{garde[m] be...}, and so on.), instead of presenting them with the full word pair \textit{garde[m] bench}. This approach could deepen our understanding of the timecourse at which compensation occurs in humans.

In future work, we aim to explore the extent to which the current findings generalize to other phonological processes. While the focus of this article on place assimilation was relatively narrow, we found some hints that the model may have knowledge about other phonological phenomena such as t-lenition and vowel reduction. Further experimentation is needed to validate this observation.

Additionally, a better replication of the original human experiments is desirable. Instead of analyzing compensation patterns in the output transcriptions of ASR models, one could train a probing model to detect the presence of a word in an assimilated sequence (e.g., the presence of the word \textit{clean} in the viable sequence \textit{clea[m] pan} versus the unviable sequence \textit{clea[m] spoon}). This setup would also allow us to probe compensation patterns in models that are entirely self-supervised and have never seen text, which are potentially more cognitively plausible than the ASR models that were used in this study.

\begin{acknowledgments}
We are grateful to Michael Hanna, Tom Lentz, Grzegorz Chrupała, Zo\"e Prins, and Katie Mudd for their assistance and insightful feedback during this project. This research is funded by the Netherlands Organisation for Scientific Research (NWO) through NWA-ORC grant NWA.1292.19.399 for `InDeep', and Gravitation grant 024.001.006 for `Language in Interaction'.
\end{acknowledgments}

\begin{multicols}{2}{
\bibliography{compling_style}
}
\end{multicols}

\newpage
\appendix

\appendixsection{Compensation patterns for a different speaker}\label{appendix:different-speaker-results}

Results for Experiment 1, 2, and 3, but with stimuli pronounced by a different American English speaker (male).

\begin{figure}[h!]
      \centering
      \includegraphics[width=0.75\linewidth]{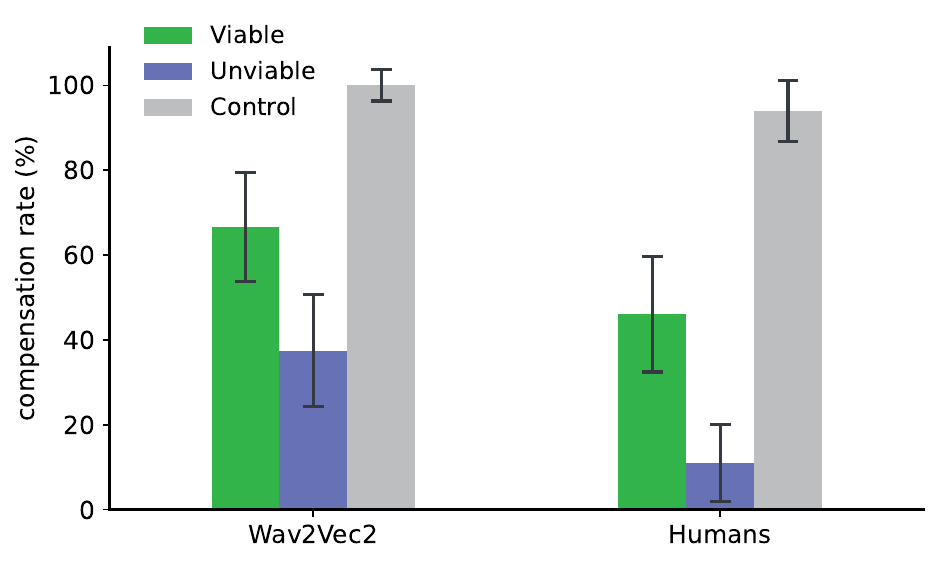}
\caption{Compensation rate (i.e., the proportion of stimuli for which the underlying consonant is transcribed) of Wav2Vec2 and humans in \textbf{viable} and \textbf{unviable} assimilation contexts (e.g., \textit{clea[m] pan} versus \textit{clea[m] spoon}, respectively). In the \textbf{control} condition, the target consonant is not assimilated (e.g., \textit{clea[n] fork}). \textit{N} = 48 for each condition. Error bars denote the 95\% Wilson confidence interval.}
\label{fig:compensation_rate_darcy_michael}
\end{figure}

\begin{figure}[h!]
      \centering
      \includegraphics[width=0.75\linewidth]{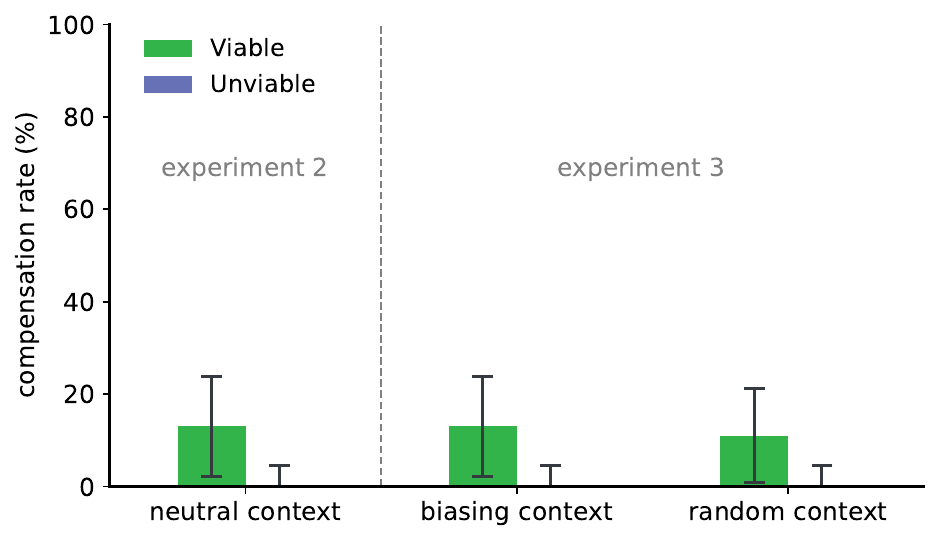}
\caption{Compensation rate (i.e., the proportion of stimuli for which the underlying consonant is transcribed) of Wav2Vec2 in \textbf{viable} and \textbf{unviable} assimilation contexts (e.g., \textit{ru[m] picks} versus \textit{ru[m] does}, respectively), with different types of preceding sentential context (\textbf{neutral context}, \textbf{biasing context}, and \textbf{random context}). \textit{N} = 38 for each condition. Error bars denote the 95\% Wilson confidence interval.}
\label{fig:compensation_rate_gaskell}
\end{figure}

\newpage
\appendixsection{Frequency of word pairs in finetuning data}\label{appendix-frequency}

\small
\centering
\begin{longtable}{l|lll|ll}
\caption{Transcriptions for stimuli by \citet{darcy2009phonological}, along with bigram frequency in the LibriSpeech finetuning data. \textit{Strict} refers to a strict regular expression match (i.e., the exact bigram occurs in the training data), \textit{loose} refers to a loose regular expression match (i.e., a close match of the word pair occurs in the training data, e.g., \textit{own planning} instead of \textit{own plan}).} \\

\hline
\textbf{Input} & \multicolumn{3}{l}{\textbf{Transcription}} & \multicolumn{2}{l}{\textbf{Frequency}} \\
\hline
 & Carrier 1 & Carrier 2 & Carrier 3 & Strict & Loose \\
\hline
\endfirsthead

\multicolumn{6}{l}%
{\tablename\ \thetable\ -- \textit{Continued from previous page}} \\
\hline
\textbf{Input} & \multicolumn{3}{l}{\textbf{Transcription}} & \multicolumn{2}{l}{\textbf{Frequency}} \\
\hline
 & Carrier 1 & Carrier 2 & Carrier 3 & Strict & Loose \\
\hline
\endhead

\hline \multicolumn{6}{l}{\textit{Continued on next page}} \\
\endfoot

\hline
\endlastfoot

clea[m] pan & clean pan & clean pan & clean pan & 0 & 1 \\
clea[m] spoon & clean spoon & clean spoon & clean spoon & 0 & 1 \\
clea[n] fork & clean fork & clean fork & clean fork & 0 & 0 \\
\midrule
ow[m] plan & own plan & own plan & own plan & 31 & 183 \\
ow[m] choice & own choice & own choice & own choice & 160 & 162 \\
ow[n] life & own life & own life & own life & 1025 & 1121 \\
\midrule
ta[m] belt & ta\textcolor{red}{m} belt & ta\textcolor{red}{m} belt & ta\textcolor{red}{m} belt & 0 & 0 \\
ta[m] shirt & ta\textcolor{red}{m}'s shirt & ta\textcolor{red}{m} shirt & ta\textcolor{red}{m}d shirt & 2 & 2 \\
ta[n] scarf & tan scarf & tan scarf & tan scarf & 0 & 0 \\
\midrule
lea[m] back & lean back & lean back & lean back & 52 & 57 \\
lea[m] shape & lea\textcolor{red}{m} shape & lea\textcolor{red}{m}s shape & lea\textcolor{red}{m} shape & 0 & 1\\
lea[n] line & lean line & lean line & lean line & 1 & 84 \\
\midrule
thi[m] packet & thin packet & thin packet & thi\textcolor{red}{m} packet & 3 & 3 \\
thi[m] leaflet & thi\textcolor{red}{m} leaflet & thin leaf let & then leaflet & 0 & 0 \\
thin notebook & thin notebook & thin notebook & the notebook & 1 & 1 \\
\midrule
grea[k] cruise & great cruise & great cruise & great cruise & 0 & 1 \\
grea[k] match & grea\textcolor{red}{k} match & great match & grea\textcolor{red}{k} match & 40 & 43 \\
grea[t] fight & great fight & great fight & great fight & 80 & 115 \\
\midrule
swee[k] cocktail & sweet cocktail & sweet cocktail & sweet cocktail & 2 & 2 \\
swee[k] liquor & swee\textcolor{red}{k} liquor & swee\textcolor{red}{k} liquor & sweet liquor & 0 & 0 \\
swee[t] chocolate & sweet chocolate & sweet chocolate & sweet chocolate & 2 & 3 \\
\midrule
fa[p] puppy & fat puppy & fat puppy & fat puppy & 1 & 1 \\
fa[p] squirrel & fa\textcolor{red}{p} squirrel & fat squirrel & fat squirrels & 1 & 1 \\
fa[t] monkey & fat monkey & fat monkey & fat monkey & 1 & 1 \\
\midrule
we[p] pants & wet pants & we\textcolor{red}{p}t pants & wet pants & 0 & 0 \\
we[p] socks & we\textcolor{red}{p}t socks & wet socks & wet socks & 1 & 1 \\
we[t] shoes & wet shoes & wet shoes & wet shoes & 11 & 11 \\
\midrule
gree[ng] cup & green cup & green cup & green cup & 0 & 5 \\
gree[ng] chair & green chair & green chair & green chair & 5 & 7 \\
gree[n] vase & green vase & green vase & green vase & 3 & 4 \\
\midrule
plai[ng] condoes & playi\textcolor{red}{ng} condoes & playi\textcolor{red}{ng} condoes & playi\textcolor{red}{ng} condoes & 0 & 0 \\
plai[ng] churches & playi\textcolor{red}{ng} churches & playi\textcolor{red}{ng} churches & playi\textcolor{red}{ng} churches & 0 & 0 \\
plai[n] chapels & plain chapels & plain chapels & plain chapels & 0 & 0 \\
\midrule
fu[ng] game & fu\textcolor{red}{ng} game & fun game & fun game & 0 & 0 \\
fu[ng] night & fu\textcolor{red}{ng} night & fu\textcolor{red}{ng} night & fu\textcolor{red}{ng} night & 0 & 0 \\
fu[n] day & fun day & fun day & fun day & 0 & 0 \\
\midrule
ma[b] brother & mad brother & mad brother & mad brother & 0 & 0 \\
ma[b] daughter & ma\textcolor{red}{pp} daughter & ma\textcolor{red}{b} daughter & ma\textcolor{red}{b} daughter & 0 & 0 \\
ma[d] mother & mad mother & mad mother & mad mother & 0 & 0 \\
\midrule
sa[b] ballet & sa\textcolor{red}{b} balet & sa\textcolor{red}{b} balet & sa\textcolor{red}{b} ballet & 0 & 0 \\
sa[b] novel & sa\textcolor{red}{b} novel & a\textcolor{red}{b}novel & sa\textcolor{red}{b} novel & 0 & 0 \\
sa[d] movie & sad movi & sad movie & sad movy & 0 & 0 \\
\midrule
ba[b] beer & bad beer & bad beer & bad beer & 11 & 11 \\
ba[b] lunch & ba\textcolor{red}{b} lunch & bad lunch & bad lunch & 0 & 0 \\
ba[d] dish & bad dish & bad dish & bad dish & 0 & 1 \\
\midrule
re[g] glasses & red glasses & red glasses & red glasses & 0 & 12 \\
re[g] lipstick & re\textcolor{red}{g} lip stickas & re\textcolor{red}{g} lipstick & re\textcolor{red}{g}lip stic & 0 & 0 \\
re[d] necklace & red necklace & red necklace & read necklace & 0 & 2 \\
\end{longtable}

\newpage
\appendixsection{Stimuli for Experiment 2 and 3}\label{appendix:lexamb-stimuli}

\begin{enumerate}
    \item \textbf{cod/cob} The chef swiftly removed the head and tail, then checked the diners’ order. They asked for the cob [poached/too late]. 
    \item \textbf{bride/bribe} The ceremony was held in June and the sunny weather added to the air of celebration. An article about the bribe [made the local paper/turned up in the local paper].
    \item \textbf{lead/leg} The council was worried about the effects on health of the old water pipes. They got the leg [covered immediately/tested immediately]. 
    \item \textbf{fad/fag} Fashions are OK, but they can be dangerous. That new fag [causes cancer/tends to cause cancer].
    \item \textbf{thud/thug} We woke up with a start from a deep sleep. A terrible thug [caught us by surprise/took us by surprise].
    \item \textbf{mud/mug} The conditions in the outback were difficult for driving. In the intense heat, the mug [cracked up completely/turned to dust]. 
    \item \textbf{road/rogue} Kate was a bit worried about the route she was taking. After a few miles, the rogue [cut across the desert/turned North across the desert].
    \item \textbf{phone/foam} The head office of the telecommunications company was empty. The manager was at the foam [packaging department/distributors].
    \item \textbf{scene/seam} We were impressed by her stylish delivery and intonation. Jane finished off the seam [beautifully/deftly].
    \item \textbf{bean/beam} She was learning about planting her allotment the hard way. Mary threw the beam [promptly on the ground/dutifully on the ground].
    \item \textbf{worn/warm} The hotel room was surprisingly shabby and the bedclothes had seen better days. It was a rather warm [blanket/duvet].
    \item \textbf{run/rum} It’s best to start the day with a burst of activity. I think a quick rum [picks you up/does you good]. 
    \item \textbf{cone/comb} His daughter had thrown the building blocks all over the place. Harry found the comb [pretty quickly/down on the floor].
    \item \textbf{turn/term} Pete was listening to the radio on the way home. Because he wasn’t concentrating, the term [passed him by completely/took him by surprise].
    \item \textbf{Dane/Dame} The debate on bacon prices drew representatives from all over Northern Europe. After the speech, the Dame [planned to leave/decided to leave].
    \item \textbf{cat/cap} It would normally turn up at feeding time, rubbing her ankles and looking hungry. Julie saw the cap [by the front door/next to the post-box].
    \item \textbf{heat/heap} The oven was switched on just before dinnertime. Nick started to heap [bowls on the stove/dinner plates on the stove].
    \item \textbf{mat/map} He was irritated by the newly uncovered dust in the doorway. Stephen put the map [back where he found it/down for a moment].
    \item \textbf{sit/sip} The children were guided to the desks in front of the blackboard. They had to sip [meekly while the party carried on/daintily while the party carried on]. 
    \item \textbf{wit/whip} His arguments were always elegant and entertaining. Michael used his whip [brilliantly/discerningly].
    \item \textbf{grit/grip} County councils have improved their safety measures for winter road conditions when bad weather is forecast. They appear to grip [motorways much better/newer roads much better].
    \item \textbf{sleet/sleep} The conditions had worsened considerably while he stood queuing outside the football ground. Four hours of intermittent sleep [brought out the worst in Mark/destroyed Mark’s good humor].
    \item \textbf{port/pork} They were obviously in the mood to get drunk. The customers had most of the pork [guzzled early on/demolished early on]. 
    \item \textbf{bait/bake} The competition on the banks of the river was well attended. The new fish bake [got tested yesterday/didn’t work very well].
    \item \textbf{late/lake} Nothing had happened for quite a while, then right at the end there was a flurry of activity. The council found the lake [growth surprising/display interesting].
    \item \textbf{dart/dark} Sarah folded the sheet of paper carefully and aimed it at the waste-bin. The dark [gradually fell/descended slowly].
    \item \textbf{bite/bike} The apple looked so appealing that he couldn’t help himself. Paul took a bike [guiltily/deliberately].
    \item \textbf{net/neck} Philip’s nephew was trying to catch shrimps in a rockpool. The little boy’s neck [got cut by the blade/nearly got cut by the blade].
    \item \textbf{wait/wake} Ben usually went straight to the bus stop after a late shift at work. He would wake [grumpily at ten o’-clock/daily at ten o’clock].
    \item \textbf{rat/rack} It’s eyes were barely visible beneath the layers of dirt. The filth from the ancient house left the rack [grimy/dirty].
    \item \textbf{line/lime} Lee preferred very abstract designs for his clothes. The T-shirt had a  lime [print on it/drawing on it].
    \item \textbf{right/ripe} With such a vast array of delicious fruit to choose from, they were spoilt for choice. They picked the ripe [berries for the pie/nectarines for the punch].
    \item \textbf{fort/fork} You have to climb up the ridge until you enter a large archway. When you reach the fork [go left/don’t turn left].
    \item \textbf{bud/bug} Patrick did everything he could to nurture his garden, but the late frost had taken its toll. When he sprayed it, the bug [curled up and died/twisted slightly].
    \item \textbf{lead/league} Second place was all they could hope to achieve. Their final game was against the league [cricket team/tennis team] 
    \item \textbf{gun/gum} There are many theories about the increase in serious crime. The role of the gum [peddlers has never been in doubt/dealers has never been in doubt].
    \item \textbf{beat/beak} Many rock songs vary enormously during a performance. Nonetheless, the beak [goes on throughout/tends to keep the tune together].
    \item \textbf{oat/oak} The traditional Scottish breakfast is often unfairly criticized. There’s no doubt that the oak [gives you vital vitamins and minerals/tastes good when it’s served properly].
\end{enumerate}

\newpage
\appendixsection{Probing results for stops}\label{appendix-probing}

\begin{figure*}[h!]
      \centering
      \includegraphics[width=0.75\linewidth]{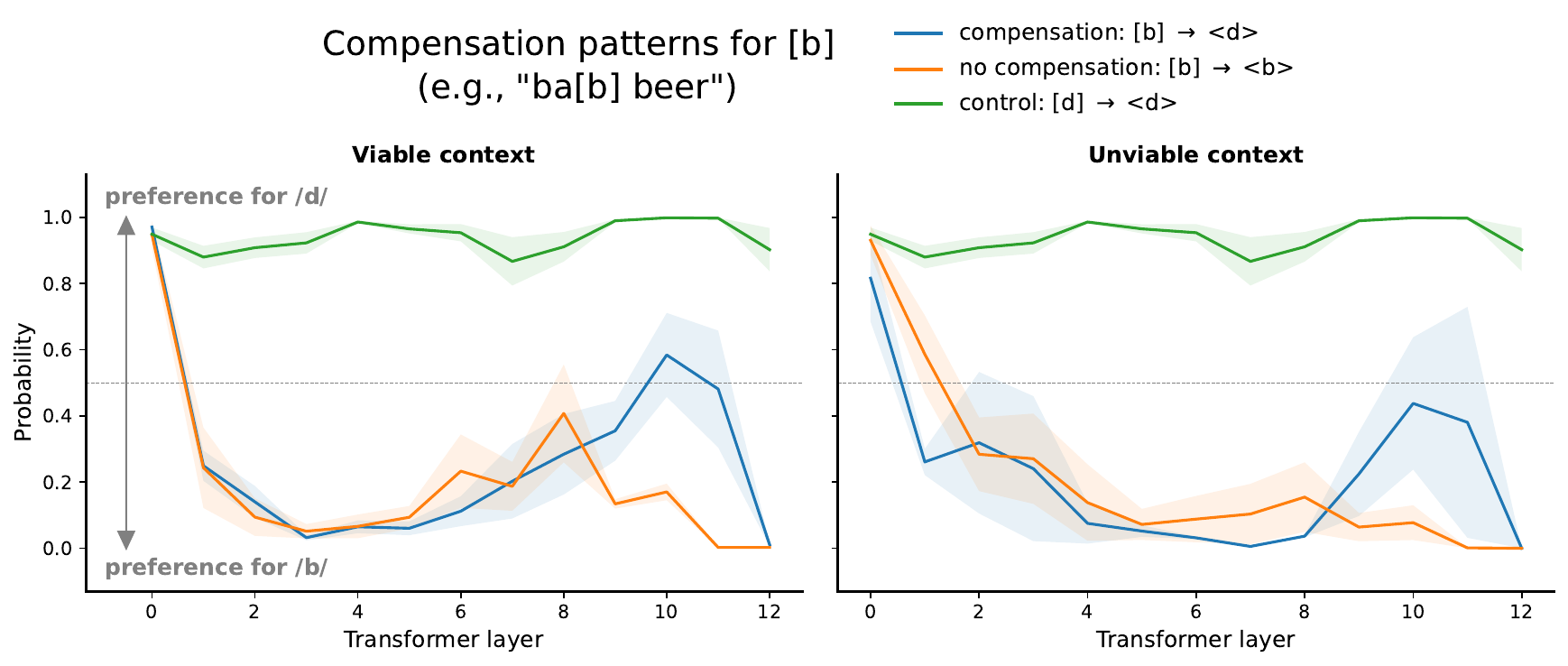}
      \includegraphics[width=0.75\linewidth]{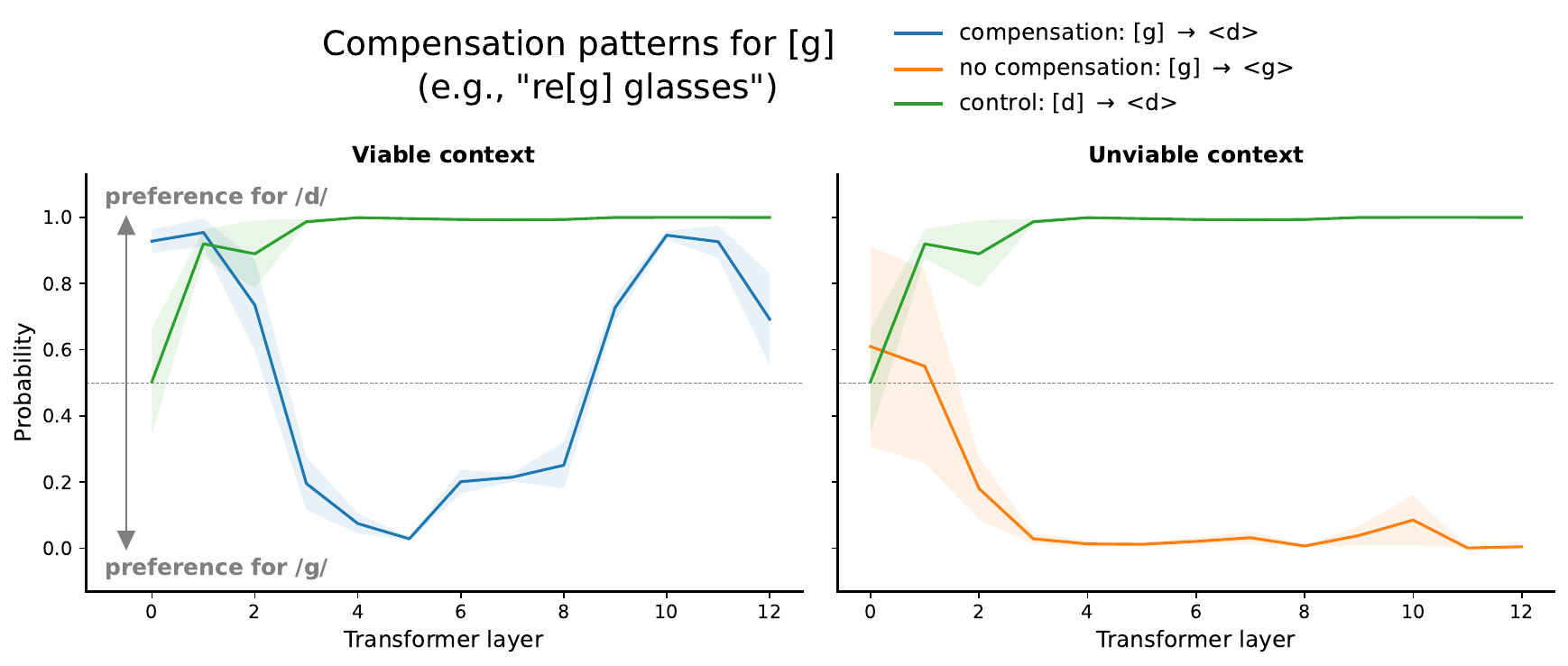}
      \includegraphics[width=0.75\linewidth]{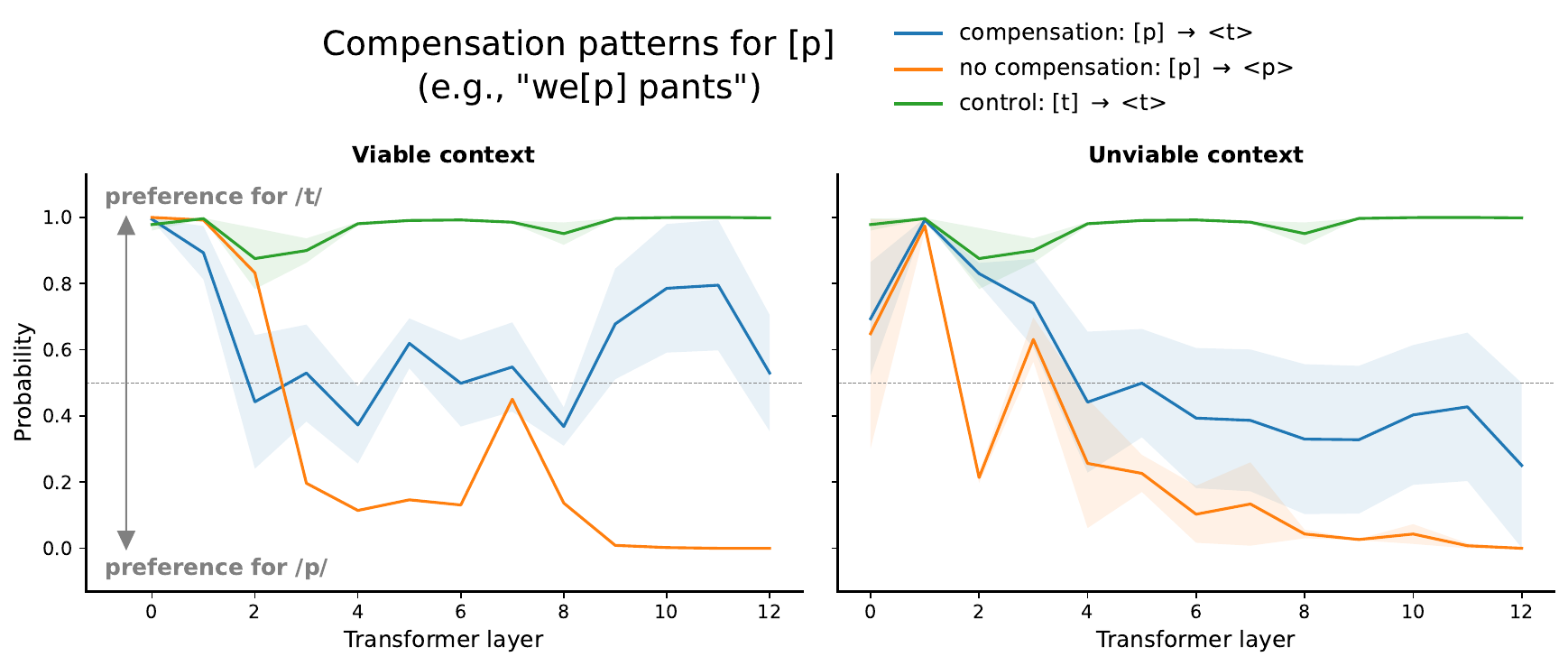}
      \includegraphics[width=0.75\linewidth]{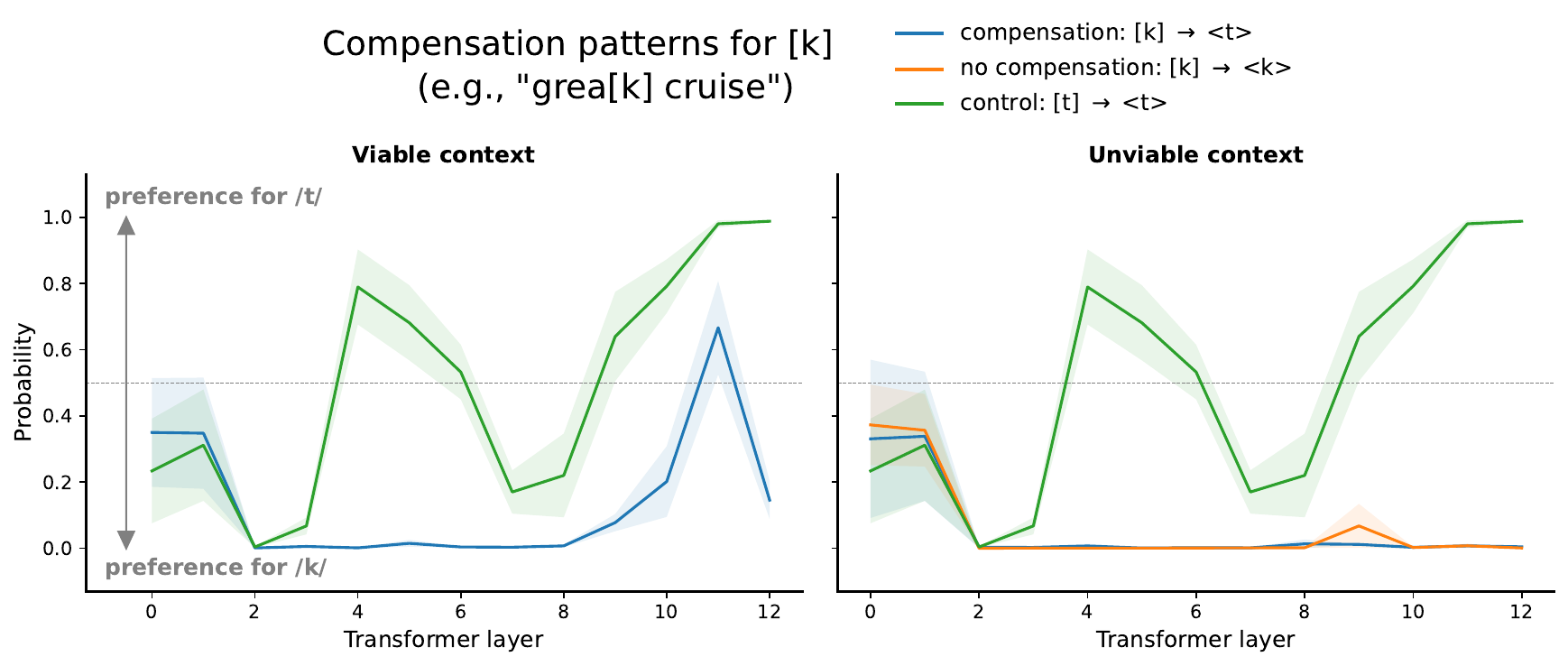}
\caption{Layerwise preference of linear probing classifiers for the underlying consonant versus the surface consonant given Wav2Vec2 representations at the position of the assimilated consonant. The three line colors indicate whether the model compensated for the assimilation in its final transcription. If one of the lines is missing, that means that the model did not show that behavior for that particular condition. Error bars denote the standard error.}
\label{fig:probing-stops}
\end{figure*}


\end{document}